\documentclass[sigconf, screen]{acmart}
\usepackage{multirow}
\AtBeginDocument{%
  }

\setcopyright{none}
\renewcommand\footnotetextcopyrightpermission[1]{}

\begin{document}

\title{InsEdit: Towards Instruction-based Visual Editing via Data-Efficient Video Diffusion Models Adaptation}

\author{Zhefan~Rao\textsuperscript{1*}, Bin~Zou\textsuperscript{2*}, Haoxuan~Che\textsuperscript{2\textdagger\textdaggerdbl}, Xuanhua~He\textsuperscript{1}, Chong~Hou~Choi\textsuperscript{2}, Yanheng~Li\textsuperscript{3}, \\ Rui~Liu\textsuperscript{2\textdagger}, Qifeng~Chen\textsuperscript{1\textdagger}}
\affiliation{%
  \institution{\textsuperscript{1} The Hong Kong University of Science and Technology}
  \city{}
  \country{}
}
\affiliation{%
  \institution{\textsuperscript{2} Celia Research HK}
  \city{}
  \country{}
}
\affiliation{%
  \institution{\textsuperscript{3} City University of Hong Kong}
  \city{}
  \country{}
}

\renewcommand{\shortauthors}{Rao et al.}

\begin{abstract}
Instruction-based video editing is a natural way to control video content with text, but adapting a video generation model into an editor usually appears data-hungry. At the same time, high-quality video editing data remains scarce. In this paper, we show that a video generation backbone can become a strong video editor without large scale video editing data. We present \textbf{InsEdit}, an instruction-based editing model built on HunyuanVideo-1.5. InsEdit combines a visual editing architecture with a video data pipeline based on \textbf{Mutual Context Attention (MCA)}, which creates aligned video pairs where edits can begin in the middle of a clip rather than only from the first frame. With only $O(100)$K video editing data, InsEdit achieves state-of-the-art results among open-source methods on our video instruction editing benchmarks. In addition, because our training recipe also includes image editing data, the final model supports image editing without any modification.

\end{abstract}


\begin{teaserfigure}
  \centering
  \includegraphics[width=\textwidth]{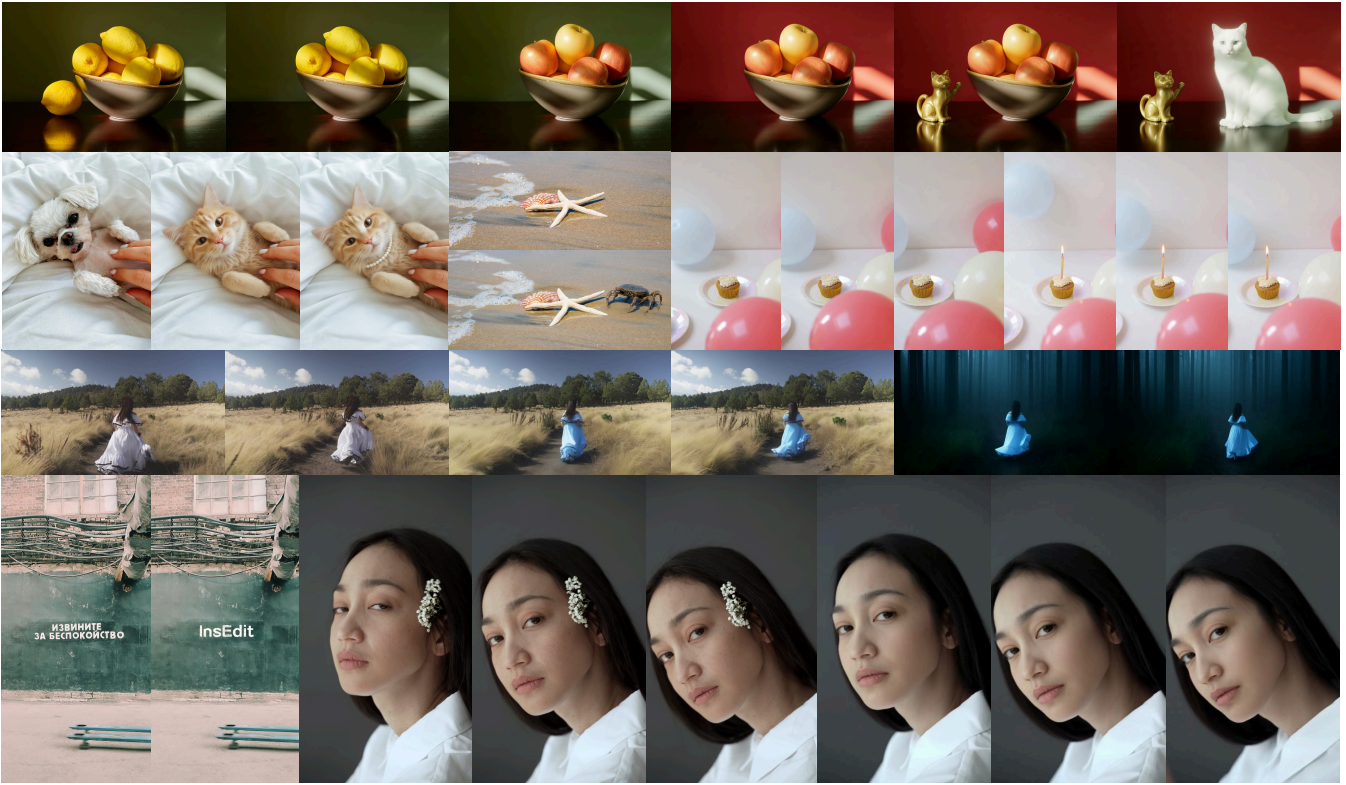}
  \caption{\textbf{Representative visual editing results of InsEdit.}}
  \vspace{0.7cm}
  \Description{A teaser figure for InsEdit showing representative visual editing results.}
  \label{fig:teaser}
\end{teaserfigure}

\maketitle
\thispagestyle{plain}
\pagestyle{plain}
\begingroup
\renewcommand{\thefootnote}{\fnsymbol{footnote}}
\setcounter{footnote}{0}
\footnotetext[1]{These authors contributed equally to this work.}
\footnotetext[2]{Corresponding author.}
\footnotetext[3]{Project leader.}
\endgroup

\clearpage
\section{Introduction}
Instruction-based video editing allows users to modify a video with natural language rather than masks, trajectories, or hand-crafted control signals. It is an attractive interface for content creation because it combines the flexibility of text with the visual quality required in modern video production. Recent progress in video generation models suggests a promising path toward this goal: a strong video generator already captures appearance, motion, and consistency over time, so it should be a good starting point for video editing. In practice, however, turning a video generation model into a reliable instruction-based editor is still difficult~\cite{Wu2025InsVie,Liao2025ICVE,Bai2025Ditto,Lin2026KiwiEdit}.

An important open question is what an efficient training path for this adaptation should look like. Recent methods often rely on large-scale video editing data, sometimes at the million-sample scale~\cite{Wu2025InsVie,openve}, yet it remains unclear how to turn a pre-trained video generator into a strong editor with a relatively modest amount of editing data. One major obstacle lies in the training data setup itself: many video editing datasets are still built by editing the first frame and propagating the result to the rest of the clip~\cite{Wu2025InsVie}. This recipe works for edits that stay active throughout the video, but it strongly biases training toward edits that start from frame one and continue to the end.

This kind of supervision does not match many real video editing requests. In practice, an edit may need to appear only after an object enters the scene, after an action starts, or during only part of a clip. These partial edits are hard to create well under the common first-frame editing-and-propagation pipeline, so current models see too few examples of one of the most important behaviors in video editing. As a result, efficient adaptation is not only a matter of model design; it also depends on whether each training sample provides enough useful editing signal.

\begin{figure}[t]
  \centering
  \includegraphics[width=0.98\columnwidth]{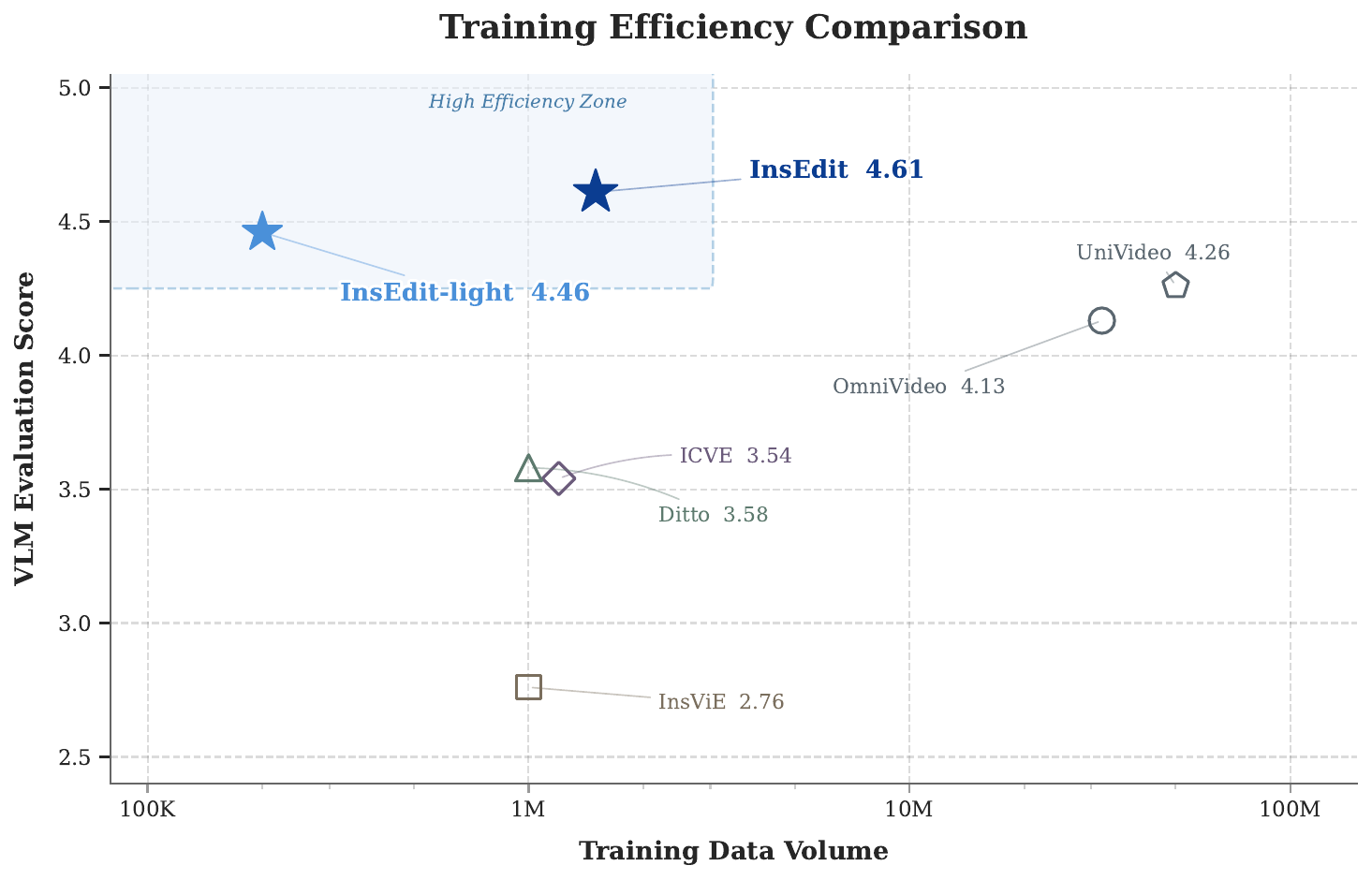}
  \vspace{-0.3cm}
  \caption{\textbf{Training efficiency of adapting video generation models into editors.}
  InsEdit reaches strong video editing performance with only $O(100)$K video editing data.}
  \Description{A single-column figure showing that InsEdit attains strong video editing performance with about one hundred thousand video editing samples.}
  \label{fig:training_efficiency}
  \vspace{-0.5cm}
\end{figure}

Our key observation is that making each training sample more informative is critical for efficient adaptation. We present \textbf{InsEdit}, an instruction-based video editing model built on HunyuanVideo-1.5, with a design that targets both sides. Prior works have shown that attention manipulation is effective for controllable generation and editing~\cite{hertz2022prompttopromptimageeditingcross,cao2023masactrltuningfreemutualselfattention,zhu2025trainingfreegeometricimageediting,xia2025dreamomni2multimodalinstructionbasedediting}. Inspired by this line of research, we develop a video data pipeline based on \textbf{Mutual Context Attention (MCA)} that jointly generates aligned source-target video pairs where the edit can begin from arbitrary points in a clip. Unlike the common first-frame-editing-and-propagation recipe, our pipeline creates more informative training pairs with naturally varying edit onset times. On the model side, we adapt the dual-stream architecture of HunyuanVideo-1.5 into an editor guided by source content, so the model can understand the editing instruction while preserving source appearance and temporal structure. Together with a simple two-stage training recipe, this gives a practical path from a video generator to a strong video editor using only $O(100)$K video editing data. As shown in Fig.~\ref{fig:training_efficiency}, this adaptation path reaches strong video editing performance with a relatively modest amount of video editing data (InsEdit corresponds to the main model in Table~\ref{tab:video_edit_main}; InsEdit-light is the lighter variant ``+ Gen Data w/ SigLIP-1'' in Table~\ref{tab:ablation_init}).

Although InsEdit is designed from a video-first view, our training recipe also includes image editing data to improve training efficiency. Because the model treats an image as a single-frame video, it naturally acquires image editing ability as well. This is not the main goal of our design, but a useful side benefit of the video-first training path.

Beyond the main model, we study which training decisions matter most for this adaptation path. We conduct systematic experiments on Stage-2 modeling choices, initialization, image-video data ratio, and prompt format, and we also construct InsEdit-Bench to evaluate diverse and challenging video editing scenarios across 13 editing categories. Experiments show that InsEdit achieves state-of-the-art results among open-source video instruction editing models on our evaluation benchmarks and gives especially strong gains on the scenarios emphasized in this paper.

Our contributions are summarized as follows:
\begin{itemize}
    \item We present \textbf{InsEdit}, which adapts HunyuanVideo-1.5 into a strong instruction-based video editor with only $O(100)$K video editing data and achieves state-of-the-art results among open-source methods on our evaluation benchmarks. Since the training recipe also incorporates image editing data, the resulting model naturally extends to image editing by treating each image as a single-frame video.
    \item We propose a video data pipeline based on Mutual Context Attention (MCA) that creates aligned video pairs where edits can begin from any point in a clip, making limited training data more informative than first-frame-propagation recipes.
    \item We construct InsEdit-Bench covering 13 editing categories and conduct systematic ablations on initialization, data ratio, and prompt format, offering practical guidance for data-efficient adaptation of video generation models to editing tasks.
\end{itemize}


\section{Related Work}

\subsection{Instruction-based Video Editing}

Instruction-based video editing is the line of work most closely related to ours. Early diffusion-based methods, such as FateZero~\cite{qi2023fatezero}, TokenFlow~\cite{geyer2023tokenflow}, FLATTEN~\cite{cong2023flatten}, RAVE~\cite{kara2024rave}, and AnyV2V~\cite{ku2024anyv2v}, mainly focus on text-guided or prompt-guided editing with handcrafted temporal constraints or tuning-free feature propagation. As video generation backbones become stronger~\cite{wu2023tune,hong2022cogvideo,kong2024hunyuanvideo,wan2025wan}, recent works move toward direct instruction-based video editing, such as InsViE-1M~\cite{Wu2025InsVie}, ICVE~\cite{Liao2025ICVE}, Ditto~\cite{Bai2025Ditto}, Kiwi-Edit~\cite{Lin2026KiwiEdit}, OpenVE-3M~\cite{openve}, Reco~\cite{Reco}, InstructVid2Vid~\cite{qin2024instructvid2vid}, VideoGrain~\cite{yang2025videograin}, VEGGIE~\cite{yu2025veggie}, Lucy-Edit~\cite{decart2025lucyedit}, and OmniVideo~\cite{tan2025omni}. These methods clearly improve edit quality and control, but most of them are still trained with supervision in which the edit is propagated from the beginning of the clip, so edits that start later in a video are less covered.

\subsection{Instruction-based Image Editing and Unified Visual Editing}

Instruction-based image editing establishes the basic source and instruction setup for language-guided visual manipulation. A representative milestone is InstructPix2Pix~\cite{Brooks2023InstructPix2Pix}, and more recent works further improve data quality, editing ability, reward modeling, instruction diversity, and benchmark coverage~\cite{nhredit,gpt-image-edit,liu2025step1x-edit,luo2025editscore,ye2025imgedit}. These studies provide a useful foundation for instruction following, but they work on static images and therefore do not address consistency over time or edits that happen only in part of a video.

An emerging line of work tries to unify visual understanding, generation, and editing across different modalities. Representative examples include UniWorld~\cite{lin2025uniworld}, DreamVE~\cite{Xia2025DreamVE}, InstructX~\cite{mou2025instructx}, UniVideo~\cite{wei2025univideo}, OmniV2V~\cite{liang2025omniv2v}, VACE~\cite{jiang2025vace}, UNIC~\cite{ye2025unicunifiedincontextvideo}, EditVerse~\cite{ju2025editverse}, UniVid~\cite{luo2025univid}, and Kling-Omni~\cite{klingteam2025klingomnitechnicalreport}. These studies suggest that image and video editing can benefit from shared backbones and shared instruction-following ability. However, existing unified frameworks focus mainly on sharing the architecture, whereas our focus is to adapt a video generation backbone and redesign the video training data itself. In our setting, image editing is an additional capability that comes from treating images as single-frame videos during training, rather than the main design target.

\subsection{Video Editing Dataset Construction}

Dataset construction is a central issue in instruction-based video editing, not just an implementation detail. Recent datasets such as Senorita-2M~\cite{zi2025se}, InsViE-1M~\cite{Wu2025InsVie}, Ditto's synthetic data~\cite{Bai2025Ditto}, and OpenVE-3M~\cite{openve} greatly expand the scale and diversity of available training data. Nevertheless, a common recipe is to edit one anchor frame, usually the first frame, and then generate the remaining frames conditionally so that the full clip stays consistent over time. In our use of Ditto and OpenVE data, we also observe that some samples built in this way show clear degradation or flickering in the first few frames, suggesting that anchor-frame editing followed by propagation can introduce instability near the beginning of a clip. While effective for producing large-scale training pairs, this recipe therefore not only covers too few edits that begin in the middle of a sequence or affect only part of it, but can also produce artifacts near the edited start of the video. Our work is closest to this line of research, but differs in directly redesigning the data construction process through an MCA-based pipeline.

\section{Method}

\subsection{Editing Architecture}

InsEdit adapts HunyuanVideo-1.5~\cite{hunyuanvideo2025}, a strong video generation model, to instruction-based editing. Each training sample contains a source image or video together with an editing instruction, and the model should apply the requested change while preserving source content and, for videos, consistency over time.

\begin{figure}[!t]
    \centering
    \includegraphics[width=0.96\columnwidth]{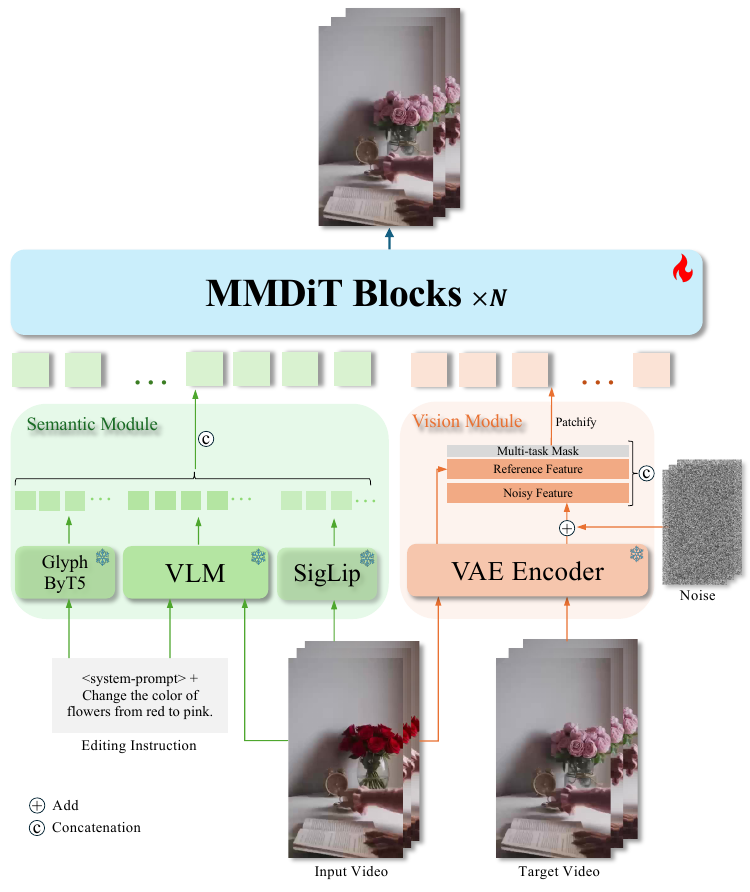}
    \caption{\textbf{Overview of the InsEdit architecture.}
    The semantic module converts the source input and editing instruction into edit-aware tokens, while the vision module denoises the target conditioned on both source features and semantic tokens.}
    \Description{A single-column pipeline figure for InsEdit. Built on HunyuanVideo-1.5, the framework contains a reference-aware semantic module that combines instruction features with source-aware cues to produce edit-aware conditioning tokens, and a vision module that combines source-conditioned visual features with noisy target latents inside shared MMDiT blocks to generate the edited target image or video.}
    \label{fig:pipeline}
\end{figure}

We keep the dual-stream design of HunyuanVideo-1.5, but make the semantic branch use the source input in addition to the instruction. As shown in Fig.~\ref{fig:pipeline}, the semantic module converts the instruction and the source input into edit-aware tokens, while the vision module reuses the original denoising stream to generate the edited target. The same design also works for images by treating an image as a single-frame video.

\paragraph{Semantic Module.}
The semantic module converts the instruction and the source input into edit-aware tokens. Following HunyuanVideo-1.5, we use three frozen encoders: Qwen2.5-VL, SigLIP, and Glyph-ByT5. The source image or video is fed into Qwen2.5-VL and SigLIP, and the resulting visual features are combined with text features from Glyph-ByT5. These features are concatenated and projected into a shared token sequence, which is injected into the MMDiT blocks.

\begin{figure*}[!t]
    \centering
    \includegraphics[width=0.96\textwidth]{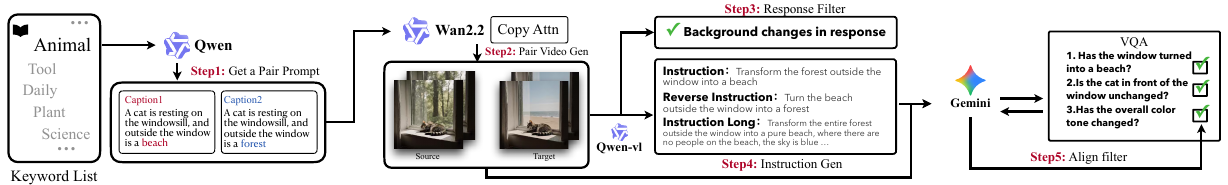}
    \caption{\textbf{Automatic pipeline for video editing data construction.}
    Starting from keyword prompts, the pipeline synthesizes aligned source-target video pairs, converts them into multiple instruction forms, and filters them with VQA-based alignment checking.}
    \Description{A pipeline diagram for automatic video editing data construction. The process starts from keyword prompts, expands them into source and target prompts, synthesizes aligned source-target video pairs with Mutual Context Attention, converts each pair into multiple instruction formats, and filters the results with VQA-based alignment verification.}
    \label{fig:data_pipeline}
\end{figure*}

\begin{figure*}[!t]
    \centering
    \includegraphics[width=0.96\textwidth]{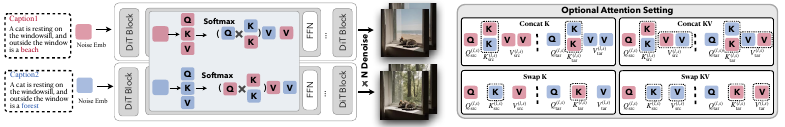}
    \caption{\textbf{Mutual Context Attention (MCA) for paired video generation.}
    Two denoising branches interact within a shared DiT backbone and exchange attention context so that the source and target videos remain aligned while still supporting diverse editing operations.}
    \Description{A diagram of the Mutual Context Attention mechanism for paired video generation. Two denoising branches for the source and target videos run inside a shared DiT backbone, and cross-branch interactions let them exchange attention context to keep the generated pair aligned while enabling different editing operations.}
    \label{fig:copyattn}
\end{figure*}

\paragraph{Vision Module.}
The vision module reuses the denoising stream of HunyuanVideo-1.5 and is initialized from its pre-trained backbone~\cite{hunyuanvideo2025}. We encode the source image or video with a frozen VAE, while the target branch starts from noisy latents. During denoising, the MMDiT blocks combine source visual features with semantic tokens to predict the edited target latent. Only the MMDiT blocks are trainable; the semantic encoders and the VAE stay frozen.


\subsection{MCA-based Video Data Construction}

Existing video editing datasets are often constructed by editing the first frame and propagating the result to the remaining frames. Although this recipe maintains temporal consistency, it biases training toward always-on edits and underrepresents cases where the change appears only in part of a video.

To reduce this mismatch, we build a video data pipeline based on \textbf{Mutual Context Attention (MCA)} that produces paired source-target videos together with matched editing instructions. The source and target branches are denoised jointly within the same DiT backbone, and at selected layers and steps MCA replaces independent self-attention with controlled cross-branch interaction, so the two branches share scene context while keeping enough freedom for the target to realize the requested change.

As shown in Fig.~\ref{fig:data_pipeline}, the pipeline has five stages: (1)~caption generation from sampled keywords via an LLM (Qwen3), (2)~paired video synthesis with MCA via a DiT generator (Wan2.2), (3)~response filtering, (4)~instruction generation via a VLM (Qwen3-VL), and (5)~multi-round VQA-based verification via a multimodal verifier (Gemini).

The paired video generation stage is the core of this pipeline because it determines whether the source and target videos stay aligned while differing only in the desired edit. MCA denoises the two branches in parallel within the same backbone and, at selected layers and denoising steps, replaces independent self-attention with controlled cross-branch interaction. In this way, MCA provides a shared scene prior while keeping branch-specific queries, which helps preserve layout, identity, and motion alignment without collapsing the intended edit difference. This design supports diverse editing operations, including object insertion and removal, local attribute modification, background replacement, motion transformation, and viewpoint change.

We formalize MCA as a branch-interaction policy over attention states. For branch $b \in \{\mathrm{src}, \mathrm{tar}\}$, let $\neg b$ denote the opposite branch. At DiT layer $l$ and denoising step $s$, the attention output of branch $b$ is defined as
\begin{equation}
\mathrm{Attn}_{b}^{(l,s)}
=
\mathrm{softmax}
\left(
\frac{
Q_{b}^{(l,s)} \left(\bar{K}_{b}^{(l,s)}\right)^\top
}{
\sqrt{d}
}
\right)
\bar{V}_{b}^{(l,s)},
\end{equation}
where $Q_{b}^{(l,s)}$ remains branch-specific, while $\bar{K}_{b}^{(l,s)}$ and $\bar{V}_{b}^{(l,s)}$ are determined by an MCA policy. We define four interaction variants plus the standard self-attention fallback:
\begingroup
\small
\begin{equation}
\begin{aligned}
\Psi_{\textsc{Self}}:~&
\bar{K}_{b}=K_{b}, \qquad\quad\;
\bar{V}_{b}=V_{b}, \\[2pt]
\Psi_{\textsc{Concat K}}:~&
\bar{K}_{b}=[K_{b};K_{\neg b}], \qquad
\bar{V}_{b}=V_{b}, \\[2pt]
\Psi_{\textsc{Concat KV}}:~&
\bar{K}_{b}=[K_{b};K_{\neg b}], \qquad
\bar{V}_{b}=[V_{b};V_{\neg b}], \\[2pt]
\Psi_{\textsc{Swap K}}:~&
\bar{K}_{b}=K_{\neg b}, \qquad\quad\;
\bar{V}_{b}=V_{b}, \\[2pt]
\Psi_{\textsc{Swap KV}}:~&
\bar{K}_{b}=K_{\neg b}, \qquad\quad\;
\bar{V}_{b}=V_{\neg b},
\end{aligned}
\end{equation}
\endgroup
where $[\cdot;\cdot]$ denotes token-wise concatenation. As illustrated in Fig.~\ref{fig:copyattn}, concat-based variants softly augment each branch's context, while swap-based variants directly replace it, providing stronger coupling at the cost of reduced branch independence.

In practice, MCA is instantiated as a task-aware schedule rather than a single fixed policy. The key insight is that early denoising mainly determines global layout and camera motion, where stronger alignment (e.g., \textsc{Swap KV}) is beneficial, while middle denoising suits softer interaction (e.g., \textsc{Concat KV}) to balance shared structure and branch-specific edits. Strong cross-branch coupling is typically disabled in late denoising to avoid texture artifacts. Detailed task-specific schedules are provided in the supplementary material.

\subsection{Details of InsEdit Data}

\begin{figure}[t]
    \centering
    \includegraphics[width=0.98\columnwidth]{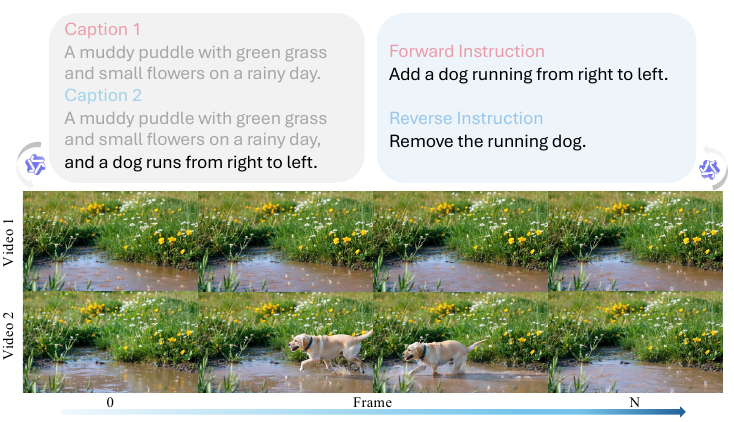}
    \caption{\textbf{Examples from the MCA-based data construction pipeline.}
    Representative video pairs illustrate temporally localized edits synthesized by our automatic pipeline, where the edited content can emerge only in intermediate video segments instead of being propagated from the first frame.}
    \Description{Representative source-target pairs produced by the MCA-based data construction pipeline, highlighting temporally localized edits that begin from intermediate segments rather than the first frame.}
    \label{fig:showcase_dataexample}
\end{figure}

Using the pipeline above, we construct about 300K paired source-target samples, where each sample consists of a source clip, a target clip, and a matched editing instruction. All video pairs are standardized to 480p resolution, 16 fps, and a duration of 3 seconds.

Unlike propagation-based datasets where every edit starts from the first frame, InsEdit Data includes pairs with naturally varying edit onset times. Because the source and target videos are jointly generated with MCA, the resulting pairs can preserve an unedited prefix or suffix, so the first or last frame may remain unchanged when the edit happens only in the middle of the clip. This temporal diversity is difficult to realize with first-frame editing followed by propagation. Fig.~\ref{fig:showcase_dataexample} shows representative examples from this pipeline.

Another advantage of this construction is that the visual quality of the data mainly depends on the strength of the underlying video generation backbone itself. In contrast, propagation-based pipelines are constrained by both the image editor used on the anchor frame and the propagation model used to spread the edit over time. Our construction therefore provides not only richer temporal edit patterns, but also a cleaner path to improving dataset quality as stronger text-to-video generators become available.

\begin{table*}[t]
    \centering
    \caption{Quantitative comparison on video instruction editing. For OpenVE-Bench~\cite{openve}, we report a subset of the original benchmark metrics due to space limitations; the complete table is provided in the supplementary material. For InsEdit-Bench, we report Overall, Instruction Compliance (IC), Temporal Visual Quality (TVQ), and Unedited Region Preservation (URP), together with inference latency.}
    \label{tab:video_edit_main}
    \resizebox{\textwidth}{!}{
    \begin{tabular}{lccccccccccc}
        \toprule
        \multicolumn{1}{c}{\multirow{2}{*}{Method}} & \multicolumn{6}{c}{OpenVE-Bench} & \multicolumn{4}{c}{InsEdit-Bench} & \multicolumn{1}{c}{\multirow{2}{*}{Latency (min) $\downarrow$}} \\
        \cmidrule(lr){2-7} \cmidrule(lr){8-11}
        & Overall $\uparrow$ & Local Add & Local Remove & Local Change & Subtitle Edit & Creative Edit
        & Overall $\uparrow$ & IC $\uparrow$ & TVQ $\uparrow$ & URP $\uparrow$ &  \\
        \midrule
        VACE-14B~\cite{jiang2025vace} & 3.01 & 1.76 & 3.99 & 2.47 & 4.41 & 2.17 & 3.08 & 2.39 & 2.64 & 4.21 & 9.08 \\
        OmniVideo~\cite{tan2025omni} & 3.66 & 2.80 & 4.52 & 3.75 & \textbf{4.95} & 1.13 & 4.13 & 4.04 & 4.04 & 4.32 & 20.58 \\
        InsViE~\cite{Wu2025InsVie} & 3.25 & 2.25 & 3.56 & 2.82 & 4.77 & 3.36 & 2.76 & 2.19 & 2.29 & 3.80 & 1.06 \\
        Lucy-Edit~\cite{decart2025lucyedit} & 3.77 & 3.92 & 3.95 & 3.93 & 4.23 & 4.19 & 3.64 & 3.24 & 3.38 & 4.30 & \textbf{0.6} \\
        ICVE~\cite{Liao2025ICVE} & 3.76 & 3.77 & 4.50 & 3.87 & 4.68 & 3.54 & 3.54 & 2.91 & 2.94 & 4.79 & 20.60 \\
        Ditto~\cite{Bai2025Ditto} & 3.44 & 2.48 & 3.53 & 2.89 & 3.69 & 4.14 & 3.58 & 3.45 & 3.52 & 3.77 & 7.33 \\
        UniVideo~\cite{wei2025univideo} & 4.21 & 4.41 & 4.46 & 4.33 & 4.56 & 4.33 & 4.26 & 4.11 & 4.07 & 4.59 & 21.5 \\
        VINO~\cite{chen2026vino} & 4.34 & 4.43 & 4.45 & 4.41 & 3.39 & 4.60 & 4.42 & 4.31 & 4.31 & 4.62 & 6.5 \\
        \midrule
        InsEdit (ours) & \textbf{4.43} & \textbf{4.78} & \textbf{4.64} & \textbf{4.71} & 4.73 & \textbf{4.66} & \textbf{4.61} & \textbf{4.50} & \textbf{4.54} & \textbf{4.80} & 1.95 \\
        \bottomrule
    \end{tabular}
    }
\end{table*}

\begin{figure*}[t]
    \centering
    \includegraphics[width=\textwidth]{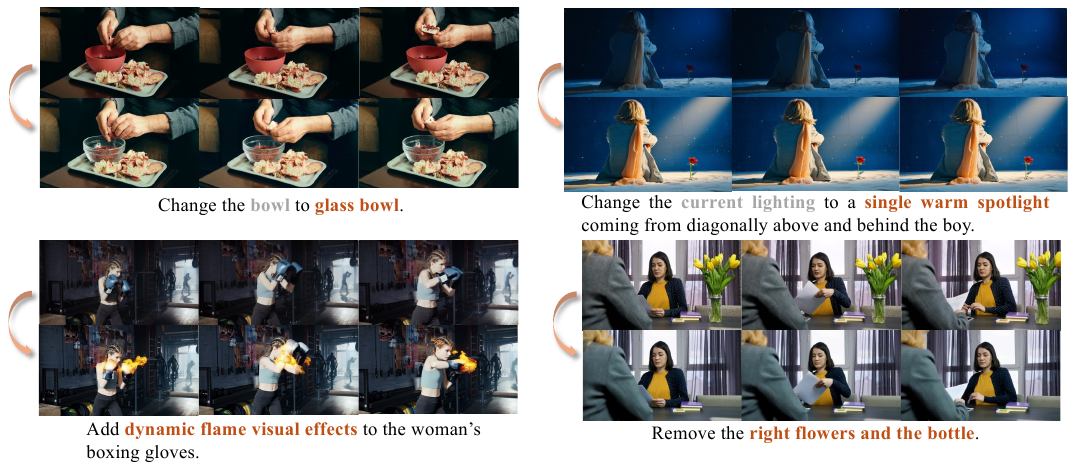}
    \caption{\textbf{Qualitative results on video instruction editing.}
    Representative examples show that InsEdit follows diverse editing instructions while preserving scene fidelity and temporal coherence.}
    \Description{A qualitative figure showing representative video instruction editing results produced by InsEdit across multiple scenarios, highlighting instruction following, temporal coherence, and preservation of unedited content.}
    \label{fig:showcase_video}
\end{figure*}

\section{Experiments}

\subsection{Implementation Details}

InsEdit is built on HunyuanVideo-1.5 and trained in two stages. Stage~1 obtains a stable initialization using three objectives—generation, VLM reconstruction, and consistency preservation—with a mixture ratio of $7{:}2{:}1$. It enables VLM visual input, uses one-frame SigLIP features, and is trained with 100K generation data. Stage~2 adapts the model to instruction-based editing. Unless otherwise specified, all results use the following Stage-2 setting: the model is initialized from the Stage~1 checkpoint and trained with $O(1)$M generation data together with $O(1)$M editing data. The generation samples are used only for the generation objective to preserve the backbone's generation ability. The default image-to-video ratio for all stages is $4{:}1$, and training prompts are randomly sampled from short instructions, long instructions, and long instructions with dense descriptions.

Although the target task is video instruction editing, our training data includes both video and image editing data. The image part is used as extra data, filtered from open-source datasets and further augmented with samples generated by Qwen-Image-Edit~\cite{wu2025qwenimagetechnicalreport}. The video part is mainly collected and filtered from OpenVE and Ditto, and is further expanded with samples generated by our MCA-based pipeline. In total, the image editing data is at the scale of $O(1)$M, while the video editing data is at the scale of $O(100)$K.

For all mainline experiments, we use AdamW with a learning rate of $2\times10^{-5}$ and train with DeepSpeed ZeRO-2. Training resolution is 480p for videos and 720p for images. During inference, we use 50 sampling steps and do not apply classifier-free guidance.

\subsection{Evaluation Setup}

Our main quantitative evaluation focuses on video instruction editing. We adopt OpenVE-Bench~\cite{openve} as the primary benchmark. In addition, to better evaluate the diverse editing scenarios emphasized in this paper, we construct InsEdit-Bench, a benchmark of 82 samples covering 13 editing categories (addition, removal, replacement, recoloring, retexturing, relocation, rescaling, background switch, weather switch, time switch, season switch, stylization, and relighting) over 80 source videos. The samples span short, long, and multi-step instructions, as well as both static and animated scenes.

For OpenVE-Bench, we follow the original benchmark protocol and report Overall together with five representative edit categories shown in the main paper: Local Add, Local Remove, Local Change, Subtitle Edit, and Creative Edit. The complete OpenVE-Bench table is provided in the supplementary material. For InsEdit-Bench, we report four summary metrics: Overall, Instruction Compliance (IC), Temporal Visual Quality (TVQ), and Unedited Region Preservation (URP). Overall measures overall editing quality, IC evaluates whether the model follows the editing instruction correctly, TVQ measures visual quality and consistency over time in the edited result, and URP measures how well non-edited regions are preserved. Following the same VLM-as-judge protocol as OpenVE-Bench, all InsEdit-Bench metrics are scored on a 1--5 scale by Qwen3-VL-32B-Instruct. In addition to editing quality, we also report inference latency measured on the same single GPU with FlashAttention-2~\cite{dao2023flashattention2}, generating 81 frames at 480p for all compared methods.

\begin{figure}[t]
    \centering
    \includegraphics[width=0.98\columnwidth]{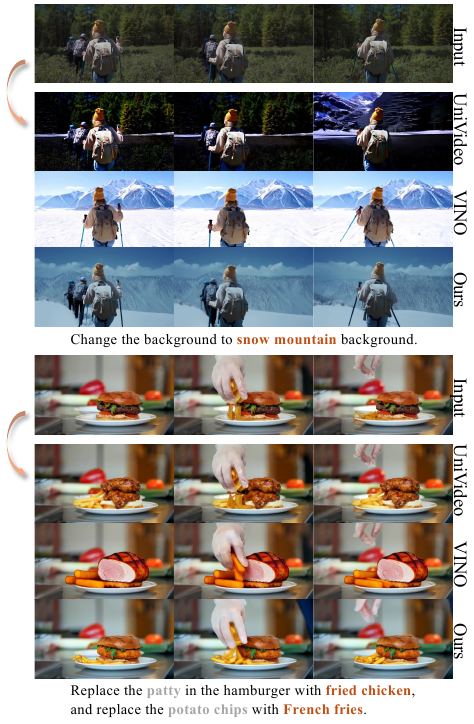}
    \caption{\textbf{Qualitative comparison with some baselines on video instruction editing.}
    Compared with UniVideo and VINO, InsEdit better follows editing instructions while preserving scene fidelity and temporal coherence.}
    \Description{A single-column qualitative comparison figure for video instruction editing, showing results from UniVideo, VINO, and InsEdit on the same editing cases.}
    \label{fig:showcase_video_compare}
\end{figure}

\subsection{Main Results on Instruction-based Video Editing}

Table~\ref{tab:video_edit_main} summarizes the main comparison against recent baselines, including VACE-14B, OmniVideo, InsViE, Lucy-Edit, ICVE, Ditto, VINO, and UniVideo. Overall, InsEdit achieves state-of-the-art performance among open-source methods on both benchmarks, showing that a video generation backbone can be effectively adapted to instruction-based video editing through a relatively modest amount of editing data when paired with the right training path.

On OpenVE-Bench, InsEdit obtains the best overall result and outperforms prior methods on most reported edit categories. The gain is especially clear on local editing metrics, where InsEdit consistently surpasses strong recent baselines such as VINO and Lucy-Edit on local addition, local removal, and local change. InsEdit also achieves the best result on creative edit while remaining competitive on subtitle edit. These results show that InsEdit improves precise control without sacrificing overall video quality.

The advantage becomes even clearer on InsEdit-Bench, which covers 13 diverse editing categories with both short and long instructions. On this benchmark, InsEdit achieves the best result on all four metrics, including overall quality, instruction compliance, temporal visual quality, and unedited-region preservation. In particular, InsEdit consistently outperforms the strongest earlier baselines, including VINO and UniVideo, indicating that the proposed training path remains effective across a wide range of video editing scenarios. In addition, among the methods with reported runtime, InsEdit also shows much lower inference latency than Ditto, suggesting that the quality gain does not come at the cost of slower inference.

Fig.~\ref{fig:showcase_video} provides representative qualitative results across diverse video editing scenarios, and Fig.~\ref{fig:showcase_video_compare} further compares InsEdit with UniVideo and VINO on challenging cases. Both figures are consistent with the quantitative results: InsEdit better follows the editing instruction while preserving scene quality and consistency over time.

\subsection{Ablation Study}

We conduct all ablations on InsEdit-Bench, since it covers diverse editing categories and includes unedited-region preservation. Our goal is to understand why this adaptation path works with relatively limited video editing data. We organize the ablations around two questions: whether Stage 2 should retain a small amount of additional generation training, and which edit-only Stage-2 recipe is most effective.

\begin{table}[t]
    \centering
    \caption{Ablation of Stage-2 modeling variants on InsEdit-Bench, including added generation data, VLM-vision input, and SigLIP temporal modeling.}
    \label{tab:ablation_init}
    \resizebox{\columnwidth}{!}{
    \begin{tabular}{lcccc}
        \toprule
        Variant & Overall $\uparrow$ & IC $\uparrow$ & TVQ $\uparrow$ & URP $\uparrow$ \\
        \midrule
        Edit-only Baseline & 4.41 & 4.24 & 4.30 & 4.70 \\
        + Gen Data w/ SigLIP-1 & \textbf{4.46} & \textbf{4.32} & \textbf{4.33} & \textbf{4.72} \\
        + Gen Data w/ SigLIP-3 & 4.30 & 4.09 & 4.17 & 4.63 \\
        + Gen Data w/ SigLIP-3 Avg & 4.32 & 4.11 & 4.21 & 4.65 \\
        + Gen Data w/o VLM Vision & 4.10 & 3.83 & 3.80 & 4.52 \\
        \bottomrule
    \end{tabular}
    }
\end{table}

\begin{table}[t]
    \centering
    \caption{Ablation of the edit-only Stage-2 recipe on InsEdit-Bench, including Stage-1 initialization, image-video training ratio, and prompt-format mixture.}
    \label{tab:ablation_stage2}
    \resizebox{\columnwidth}{!}{
    \begin{tabular}{clcccc}
        \toprule
        Group & Variant & Overall $\uparrow$ & IC $\uparrow$ & TVQ $\uparrow$ & URP $\uparrow$ \\
        \midrule
        \textit{Baseline} & Edit-only Baseline & \textbf{4.41} & \textbf{4.24} & \textbf{4.30} & \textbf{4.70} \\
        \midrule
        \multirow{3}{*}{\textit{Initialization}} & w/o Stage 1 Init & 4.15 & 3.94 & 3.95 & 4.57 \\
         & w/o Stage 1 Consistency Init & 4.26 & 3.99 & 4.04 & 4.74 \\
         & w/o Stage 1 VLM-Recon Init & 4.29 & 4.09 & 4.15 & 4.63 \\
        \midrule
        \multirow{2}{*}{\textit{Data Ratio}} & Image:Video = 1:1 & 4.34 & 4.15 & 4.16 & 4.71 \\
         & Image:Video = 1:4 & 4.27 & 4.07 & 4.12 & 4.62 \\
        \midrule
        \multirow{3}{*}{\textit{Prompt Format}} & Short Instruction only & 4.14 & 3.87 & 4.00 & 4.55 \\
         & Long Instruction only & 4.19 & 3.90 & 3.99 & 4.67 \\
         & Short/Long Mix & 4.21 & 3.88 & 4.02 & 4.72 \\
        \bottomrule
    \end{tabular}
    }
\end{table}

\paragraph{Part I: Stage-2 modeling variants.}
Table~\ref{tab:ablation_init} studies whether Stage 2 should retain a small amount of additional generation training and how the semantic module should be designed in this setting. \textit{Edit-only Baseline} is trained on 100K editing samples only. We then add 10K generation samples in Stage 2 while using them only for the generation objective, yielding \textit{+ Gen Data w/ SigLIP-1}. Starting from this setting, we further vary the semantic design. \textit{+ Gen Data w/ SigLIP-3} replaces the default single-frame SigLIP input with three-frame features, \textit{+ Gen Data w/ SigLIP-3 Avg} averages these three-frame features before injection, and \textit{+ Gen Data w/o VLM Vision} removes visual input from the VLM semantic module and keeps only the text instruction.

The results show that the simplest setting with added generation data is the most effective. Compared with \textit{Edit-only Baseline}, \textit{+ Gen Data w/ SigLIP-1} gives the best overall balance across all four metrics, indicating that retaining a small amount of generation training in Stage 2 is beneficial for instruction editing, likely because it helps preserve the backbone's generation prior during adaptation. In contrast, replacing the default single-frame SigLIP input with three-frame features leads to a clear performance drop, and averaging the three-frame features only partly reduces this drop. Finally, removing VLM vision causes the largest decline among all variants, showing that visual input in the semantic module remains crucial even when generation training is retained.

\paragraph{Part II: edit-only Stage 2.}
Table~\ref{tab:ablation_stage2} focuses on the pure edit-only Stage-2 branch, i.e., without added generation data. The purpose of this part is to answer three practical questions for the default edit-only recipe: whether Stage-1 initialization is necessary and, if so, which initialization is most useful; what image-to-video ratio should be used during Stage 2; and what prompt mixture should be adopted for training. We use \textit{Edit-only Baseline} as the anchor setting throughout. Unless otherwise specified, this baseline uses the full Stage-1 initialization, an image-to-video ratio of $4{:}1$, and a mixed prompt recipe consisting of short instructions, long instructions, and long instructions with dense descriptions.

The \textit{Initialization} group shows that Stage-1 transfer remains important even for a purely edit-only Stage 2. Training Stage 2 from scratch leads to a clear drop across all four metrics, confirming that edit-only adaptation still relies on a strong generation starting point. Among partial initializations, removing VLM reconstruction is consistently better than removing consistency preservation on Overall, IC, and TVQ, whereas removing consistency preservation yields the strongest URP. Overall, the full Stage-1 initialization remains the strongest default, and the results suggest that pretraining for consistency is especially important for later editing quality.

The \textit{Data Ratio} group indicates that image training data should be kept at a relatively high proportion during Stage 2. The default $4{:}1$ image-to-video ratio remains the strongest overall setting, while reducing the ratio to $1{:}1$ gives the closest alternative and even slightly improves preservation. In contrast, moving to the video-heavy $1{:}4$ setting causes a consistent drop across all main metrics. We interpret this pattern as follows: image data mainly strengthens spatial editing abilities, such as appearance control, local structure preservation, and instruction following, whereas video data mainly supplies the temporal supervision needed for motion and consistency over time. Under this view, a higher image ratio is not only effective but also practical, because image editing data is much cheaper to collect than video editing data. The result therefore points to a favorable training recipe: use abundant image data to build strong spatial editing ability, and use a smaller amount of video data to provide the temporal component that images cannot offer.

The \textit{Prompt Format} group compares three reduced prompt recipes against the baseline's default mixture of short instructions, long instructions, and long instructions with dense descriptions. Using only short instructions or only long instructions both lead to clear drops, confirming that prompt diversity matters. \textit{Short/Long Mix} recovers part of the gap but still falls behind the full three-way mixture, showing that long instructions with dense descriptions provide additional training signal that benefits editing quality, particularly for unedited-region preservation.

\subsection{Image Editing as an Additional Capability}

Image editing is not treated here as a separate generalization setting. Instead, it is an additional capability that comes from the training recipe itself. To improve training efficiency in Stage 2, we already include large-scale image editing data together with video editing data, and the model handles images by treating them as single-frame videos. We therefore also evaluate the final model on image instruction editing. Table~\ref{tab:image_edit_main} reports results on GEdit~\cite{liu2025step1x-edit}. We compare InsEdit with strong proprietary systems and recent open-source baselines, including UniWorld-V1~\cite{lin2025uniworld}, OmniGen2~\cite{wu2025omnigen2}, FLUX.1 Kontext [dev]~\cite{batifol2025fluxkontext}, BAGEL~\cite{deng2025bagel}, Step1X-EditV1.1~\cite{liu2025step1x-edit}, and VINO~\cite{chen2026vino}.

\begin{table}[t]
    \centering
    \caption{Quantitative comparison on GEdit~\cite{liu2025step1x-edit}. We report the three average metrics used by the benchmark.}
    \label{tab:image_edit_main}
    \resizebox{0.9\columnwidth}{!}{
    \begin{tabular}{lccc}
        \toprule
        Method & G\_SC $\uparrow$ & G\_PQ $\uparrow$ & G\_O $\uparrow$ \\
        \midrule
        Gemini2.5 & 7.48 & 8.30 & 7.17 \\
        GPT4o & 8.06 & 7.80 & 7.48 \\
        Seedream4 & 8.33 & 8.00 & 7.72 \\
        \midrule
        UniWorld-V1~\cite{lin2025uniworld} & 5.04 & 7.56 & 4.98 \\
        OmniGen2~\cite{wu2025omnigen2} & 6.79 & 6.68 & 6.18 \\
        Flux-Kontext-Dev~\cite{batifol2025fluxkontext} & 7.23 & 7.28 & 6.53 \\
        Bagel~\cite{deng2025bagel} & 7.52 & 6.69 & 6.54 \\
        Step1x-EditV1.1~\cite{liu2025step1x-edit} & \textbf{7.60} & 7.29 & 6.87 \\
        VINO~\cite{chen2026vino} & 7.26 & 7.71 & \textbf{6.88} \\
        \midrule
        InsEdit (ours) & 6.98 & \textbf{7.77} & 6.72 \\
        \bottomrule
    \end{tabular}
    }
\end{table}

The results show that InsEdit, despite not being designed as an image-first editing model, achieves the best perceptual quality (G\_PQ) among open-source methods while remaining competitive on semantic consistency (G\_SC) and overall score (G\_O). InsEdit still trails dedicated image editing models such as Step1X-Edit and VINO on G\_SC, which is expected given that our method is optimized primarily for video editing rather than image-specific editing objectives. Nevertheless, the results show that adding image data for efficient training also leaves the final video-first model with a practically useful image editing capability.

%

\section{Conclusion}

In this paper, we presented InsEdit as an efficient training path for turning a video generation backbone into an instruction-based editor. The key idea is to adapt HunyuanVideo-1.5 with a source-guided editing architecture while making the training data more informative through a video data pipeline based on Mutual Context Attention (MCA). This combination lets the model learn video editing from a relatively modest amount of editing data while staying better matched to real video editing scenarios.

Experiments validate the effectiveness of this design. With only $O(100)$K video editing data, InsEdit establishes state-of-the-art performance among open-source methods on video instruction editing benchmarks, including strong results on InsEdit-Bench across 13 diverse editing categories. Because our training recipe also uses image data, the final model additionally supports image instruction editing by treating images as single-frame videos.

Our ablation study reveals several practical findings for efficient video-to-editor adaptation: (1)~retaining a small amount of generation data in Stage~2 helps preserve the backbone's generation prior; (2)~a high image-to-video ratio (4:1) is both effective and cost-efficient, as image data strengthens spatial editing ability while video data supplies the temporal component; (3)~prompt diversity during training, particularly long instructions with dense descriptions, meaningfully improves editing quality; and (4)~Stage-1 pretraining for consistency is especially important for downstream editing performance.

A current limitation is that InsEdit still focuses on short-clip, language-driven 2D editing. We observe failure cases on edits requiring precise spatial control, multi-object relational reasoning, and long-video temporal consistency; representative examples are provided in the supplementary material. Promising future directions include extending to longer videos with multi-scene structure, incorporating spatial control signals such as masks or keypoints alongside language instructions, and scaling the MCA-based data pipeline with stronger video generation backbones to further improve data quality. We hope InsEdit can serve as a useful starting point for future work on video-first instruction-based visual editing.


\clearpage
\bibliographystyle{ACM-Reference-Format}
\bibliography{main}

\clearpage
\appendix
\section{MCA Schedule Details}

In the main paper, we describe MCA as a general framework with four atomic
interaction variants: \textsc{Concat K}, \textsc{Concat KV}, \textsc{Swap K},
and \textsc{Swap KV}. In practice, our data construction pipeline instantiates
this framework using task-specific schedules, because different editing
categories require different trade-offs between pair alignment and target
editability.

These variants provide different levels of coupling between the two branches and
can be viewed as a general design space for balancing alignment and editability.
\textsc{Concat KV} is a soft sharing policy that encourages semantically
corresponding elements in the two videos to remain similar even when they
appear in different states. In practice, it is useful for edits such as motion
or viewpoint change, where the same subject should preserve identity
consistency while allowing state differences. \textsc{Concat K} also promotes
correspondence between the two branches, but in a more state-consistent manner,
making it suitable for appearance-oriented edits such as color or material
modification, where object geometry should stay stable while visual content
changes. In contrast, \textsc{Swap KV} is the strongest interaction: it
effectively forces one branch to decode under the other branch's context, which
is particularly useful in early denoising for tightly locking camera motion and
coarse scene structure. However, our pilot study and qualitative inspection
show that using \textsc{Swap KV} in late denoising often introduces ghosting
and texture artifacts. \textsc{Swap K} provides a milder structural anchor and
is often effective for aligning the non-edited context in local-object or
background-related edits.

MCA should not be activated uniformly across the whole denoising process. Let
$\mathcal{L}^{\text{early}}$ and $\mathcal{L}^{\text{mid}}$ denote the shallow
and middle DiT layer groups, and let $\mathcal{S}^{\text{early}}$,
$\mathcal{S}^{\text{mid}}$, and $\mathcal{S}^{\text{late}}$ denote the early,
middle, and late denoising stages, respectively. In our current implementation,
MCA is instantiated as a task-aware schedule rather than a single fixed policy.
The underlying intuition is that early denoising mainly determines global
layout, camera trajectory, and coarse motion, where stronger alignment is
beneficial, whereas middle denoising is more suitable for balancing shared
structure and branch-specific edits. Late denoising primarily refines texture
and local appearance, so overly strong cross-branch coupling at this stage
tends to hurt visual quality.

\subsubsection{Task-Specific MCA Schedules}
\label{app:mca_schedule}

\begin{itemize}
    \item \textbf{Object insertion and removal.} We use an asymmetric schedule:
    one branch keeps standard self-attention to preserve natural video
    generation, while the other branch uses \textsc{Swap KV} in the early
    denoising stage, then switches to \textsc{Concat KV} in the middle stage,
    and finally returns to \textsc{Self} in the late stage. This design is
    motivated by the fact that insertion and removal often involve relatively
    large semantic changes. Strong early alignment helps preserve unchanged
    content, while releasing the constraint later allows the edited object to
    appear or disappear naturally.

    \item \textbf{Local object modification.} We apply \textsc{Swap KV} to both
    branches in the early stage and switch to \textsc{Concat KV} in the middle
    stage. This combination first aligns the non-edited context and coarse
    structure, and then allows the edited local region to deviate smoothly
    without breaking overall scene coherence.

    \item \textbf{Background replacement.} We use \textsc{Swap K} more heavily,
    mainly in shallow-to-middle layers during the early and middle denoising
    stages. This policy is effective when the scene background should stay
    structurally aligned while the foreground subject remains editable.

    \item \textbf{Color and material modification.} We first use
    \textsc{Swap KV} and \textsc{Concat KV} in the early stage to lock geometry
    and instance identity, and then rely primarily on \textsc{Concat K} in the
    middle stage. This schedule keeps the object shape and spatial extent
    stable while allowing its appearance attributes, such as color or texture,
    to change.

    \item \textbf{Motion and viewpoint transformation.} We use
    \textsc{Concat KV} throughout the selected layers and denoising steps. This
    soft sharing policy is particularly suitable when semantically
    corresponding subjects should remain consistent in identity while their
    motion state or camera relation changes over time.
\end{itemize}

Overall, these task-specific schedules are different instantiations of the same
MCA framework. A practically effective default pattern is to use stronger
swap-based interaction in early denoising to secure coarse alignment, followed
by concat-based interaction in the middle stage to preserve editability, and to
reduce strong cross-branch coupling in late denoising to avoid artifacts.

\section{More Comprehensive Results on OpenVE-Bench}

Table~\ref{tab:openve_bench_full} reports a more comprehensive category-level
breakdown on OpenVE-Bench~\cite{openve}. Compared with the main paper, we
include all edit categories defined by the benchmark and list the corresponding
baseline results together with InsEdit.

\begin{table*}[t]
    \centering
    \caption{More comprehensive category-level results on
    OpenVE-Bench~\cite{openve}.}
    \label{tab:openve_bench_full}
    \resizebox{\textwidth}{!}{
    \begin{tabular}{lccccccccc}
        \toprule
        Method & Overall $\uparrow$ & Local Add & Local Remove &
        Local Change & Background Change & Global Style & Subtitle Edit &
        Creative Edit & Camera Edit \\
        \midrule
        VACE-14B~\cite{jiang2025vace} & 3.01 & 1.76 & 3.99 & 2.47 & 2.81 & 3.46 & 4.41 & 2.17 & 3.09 \\
        OmniVideo~\cite{tan2025omni} & 3.66 & 2.80 & 4.52 & 3.75 & 4.11 & 3.41 & 4.95 & 1.13 & 3.62 \\
        InsViE~\cite{Wu2025InsVie} & 3.25 & 2.25 & 3.56 & 2.82 & 2.68 & 3.63 & 4.77 & 3.36 & 3.61 \\
        Lucy-Edit~\cite{decart2025lucyedit} & 3.77 & 3.92 & 3.95 & 3.93 & 3.25 & 3.64 & 4.23 & 4.19 & 3.54 \\
        ICVE~\cite{Liao2025ICVE} & 3.76 & 3.77 & 4.50 & 3.87 & 3.51 & 3.87 & 4.68 & 3.54 & 2.84 \\
        Ditto~\cite{Bai2025Ditto} & 3.44 & 2.48 & 3.53 & 2.89 & 3.52 & 4.48 & 3.69 & 4.14 & 3.33 \\
        OpenVE-Edit~\cite{openve} & 3.89 & 3.41 & 3.50 & 3.80 & 4.10 & 4.24 & 3.98 & 3.71 & 3.25 \\
        VINO~\cite{chen2026vino} & 4.34 & 4.43 & 4.45 & 4.41 & 4.46 & 4.78 & 3.39 & 4.60 & 4.08 \\
        \midrule
        InsEdit (ours) & 4.43 & 4.78 & 4.64 & 4.71 & 3.99 & 4.20 & 4.73 & 4.66 & 3.62 \\
        \bottomrule
    \end{tabular}
    }
\end{table*}

\section{More Qualitative Results of Image Editing on GEdit}

\begin{figure*}[t]
    \centering
    \includegraphics[width=0.98\textwidth,height=0.82\textheight,keepaspectratio]{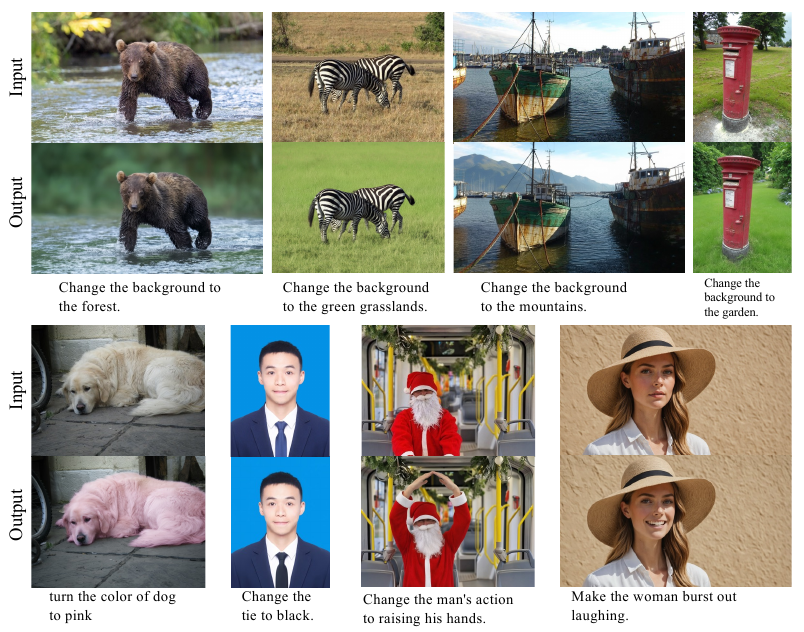}
    \caption{\textbf{Additional qualitative results on GEdit.}
    Supplementary examples illustrate the image editing performance of InsEdit
    on the GEdit benchmark.}
    \Description{A supplementary qualitative figure showing additional image
    editing results of InsEdit on the GEdit benchmark.}
    \label{fig:supp_img_showcase}
\end{figure*}

Figure~\ref{fig:supp_img_showcase} presents additional qualitative image
editing results on GEdit. The showcased examples cover a diverse set of
editing instructions, including appearance transformation, object-level
manipulation, and compositional modification. These cases further demonstrate
that InsEdit can follow user instructions faithfully while preserving the
overall scene structure and visual consistency of the original image.

\section{More Qualitative Results of Video Editing on InsEdit-Bench}

\begin{figure*}[p]
    \centering
    \includegraphics[width=0.98\textwidth,height=0.92\textheight,keepaspectratio]{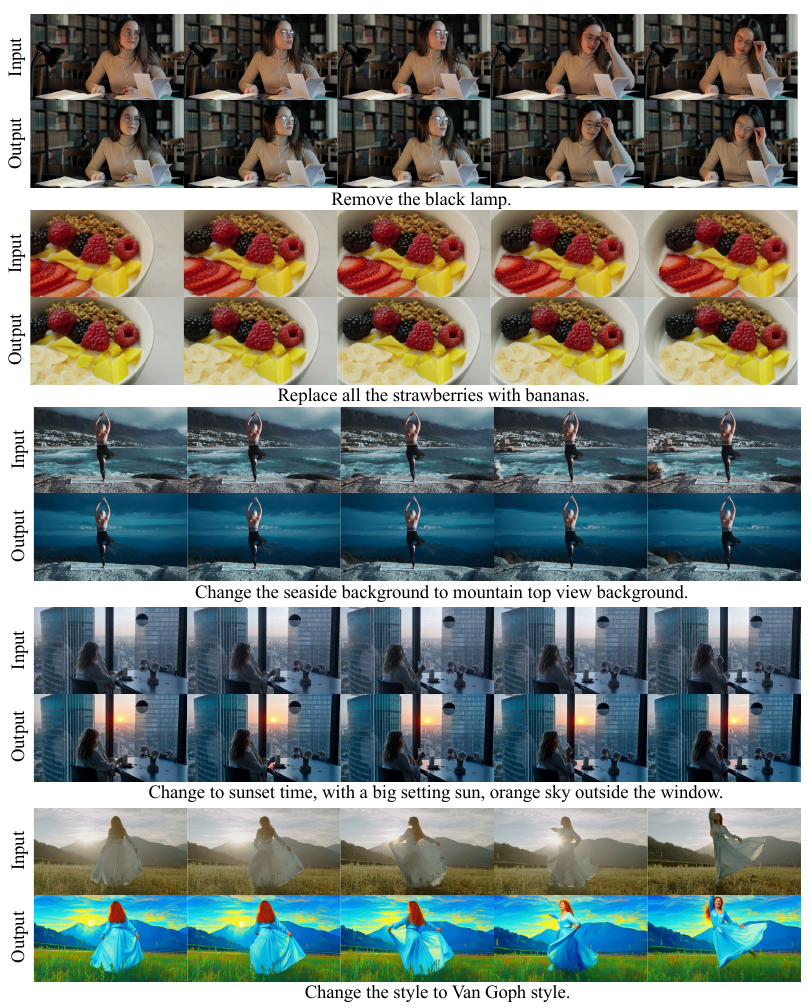}
    \caption{\textbf{Additional qualitative video editing results.}
    More supplementary examples illustrate the editing performance of InsEdit
    across diverse instructions and scenes.}
    \Description{A supplementary qualitative figure showing additional video
    editing results for InsEdit across multiple editing scenarios.}
    \label{fig:supp_video_showcase}
\end{figure*}

Figure~\ref{fig:supp_video_showcase} presents more qualitative results on
InsEdit-Bench beyond the representative cases shown in the main paper. The
examples span diverse editing categories and scene dynamics, illustrating that
InsEdit can follow challenging video editing instructions while maintaining
temporal coherence and preserving unedited content whenever possible.

\end{document}